\documentclass{IEEEtran}
\usepackage{cite}
\usepackage{amsmath,amssymb,amsfonts}
\usepackage{algorithmic}
\usepackage{graphicx}
\usepackage{textcomp}
\usepackage{multirow}
\usepackage{hyperref}
\usepackage{bm}
\usepackage{xcolor}
\usepackage{adjustbox}
\usepackage{multirow}
\usepackage{makecell} 
\usepackage[multiple]{footmisc}
\usepackage{tikz}
\usepackage{lipsum}

\def\BibTeX{{\rm B\kern-.05em{\sc i\kern-.025em b}\kern-.08em
    T\kern-.1667em\lower.7ex\hbox{E}\kern-.125emX}}
\begin{document}
\title{Mimic and Fool: A Task Agnostic Adversarial Attack}
\author{Akshay Chaturvedi$^{*}$, and Utpal Garain, \IEEEmembership{Member, IEEE}
\thanks{$^{*}$ Corresponding author. Email: akshay91.isi@gmail.com }
\thanks{Akshay Chaturvedi, and Utpal Garain are with Computer Vision and Pattern Recognition Unit, Indian Statistical Institute, Kolkata 700108, India.}}
\newcommand\copyrighttext{%
  \footnotesize \textcopyright 2020 IEEE. Personal use of this material is permitted.
  Permission from IEEE must be obtained for all other uses, in any current or future
  media, including reprinting/republishing this material for advertising or promotional
  purposes, creating new collective works, for resale or redistribution to servers or
  lists, or reuse of any copyrighted component of this work in other works.
  DOI: {10.1109/TNNLS.2020.2984972}} % DOI to be added here
\newcommand\copyrightnotice{%
\begin{tikzpicture}[remember picture,overlay]
\node[anchor=south,yshift=8pt] at (current page.south) {\fbox{\parbox{\dimexpr\textwidth-\fboxsep-\fboxrule\relax}{\copyrighttext}}};
\end{tikzpicture}%
}
\maketitle
\copyrightnotice

\begin{abstract}

At present, adversarial attacks are designed in a task-specific fashion. However, for downstream computer vision tasks such as image captioning, image segmentation etc., the current deep learning systems use an image classifier like VGG16, ResNet50, Inception-v3 etc. as a feature extractor. Keeping this in mind, we propose Mimic and Fool, a task agnostic adversarial attack. Given a feature extractor, the proposed attack finds an adversarial image which can mimic the image feature of the original image. This ensures that the two images give the same (or similar) output regardless of the task. We randomly select 1000 MSCOCO validation images for experimentation. We perform experiments on two image captioning models, Show and Tell, Show Attend and Tell and one VQA model, namely, end-to-end neural module network (N2NMN). The proposed attack achieves success rate of 74.0\%, 81.0\% and 87.1\% for Show and Tell, Show Attend and Tell and N2NMN respectively. We also propose a slight modification to our attack to generate natural-looking adversarial images. In addition, we also show the applicability of the proposed attack for invertible architecture. Since Mimic and Fool only requires information about the feature extractor of the model, it can be considered as a gray-box attack.
\end{abstract}


\section{Introduction}
\label{sec:introduction}
\begin{figure}[h!]
\begin{center}
   \includegraphics[width=0.8\linewidth,keepaspectratio]{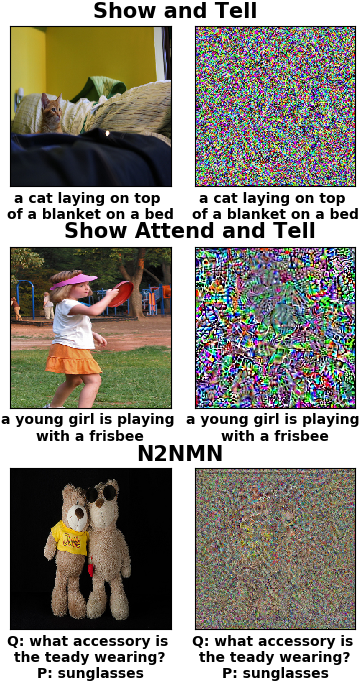}
\end{center}
   \caption{Examples of Mimic and Fool. The first two rows show the original and adversarial images along with the predicted captions by Show and Tell and Show Attend and Tell respectively. The last row shows original and adversarial image for N2NMN (Q, P denote the question and the predicted answer respectively).}
\label{fig:main}
\end{figure}
Adversarial attacks have shed light on the vulnerability of several state-of-the-art deep learning systems across varied tasks such as image classification, object detection, image segmentation etc. \cite{shapeshifter,43405,KurakinGB16,Xie2017Adversarial}. Recently, adversarial attacks were also proposed for multimodal tasks involving vision and language like image captioning and visual question answering (VQA) \cite{P18-1241,Xu-2018-CVPR}. Usually, these attacks fall under two categories: white-box and black-box. In white-box attack, the adversary has complete information about the model and its parameters. Whereas in black-box attack, the adversary has no information about the model that it wants to attack. Black-box attacks \cite{blackbox} are possible due to the \emph{transferability} phenomenon of adversarial examples. Liu \emph{et al.} \cite{transfer} show that the adversarial examples designed for one image classification model can be transferred successfully to other classification models as well. Similarly, Xu \emph{et al.} \cite{Xu-2018-CVPR} show transferability of adversarial images between two state-of-the-art VQA models. Very recently, Shi \emph{et al.}~\cite{curlswhey} improved black-box attack performance for image classification by allowing for more diverse search trajectories and squeezing redundant noise.  However, the present-day adversarial attacks are task-specific in nature since a task-specific adversarial loss function is optimized to generate adversarial examples.

 On the other hand, the current deep learning systems use output from intermediate layers of convolutional neural network (CNN) based image classification models (e.g. ResNet50 \cite{He16}, VGG16 \cite{Simonyan14c}, Inception-v3 \cite{43022} etc.) as a feature for the input image. The rationale behind this approach is that the discriminative features learned by these classifiers are useful for other vision tasks as well. Hence, it is more beneficial to use these features instead of learning them \emph{from scratch}. As a result, the aforementioned image classifiers function as \emph{feature extractors}. Some deep learning systems also \emph{fine-tune} the parameters of the \emph{feature extractors} during training to make the image feature more suitable for the task in hand. However, fine-tuning is usually done if large amount of training data is available. Although using deep CNN-based image features give significant advantage to the present-day models, they have their own set of drawbacks. CNN-based feature extractors are known to be \emph{non-invertible} \cite{DosovitskiyB16,MahendranV15}. Mahendran and Vedaldi \cite{MahendranV15} show that AlexNet \cite{NIPS2012_4824} maps multiple images to the same $1000$-dimensional logits. These images are thus \emph{indistinguishable} from the viewpoint of the last fully connected layer of AlexNet.
 
 \subsection{Motivation: Agnosticism in Adversarial Attacks}
 % Talk about black box attacks and UAP.
 The main goal of this paper is to introduce the notion of a task-agnostic attack. If such an attack were possible, it will shed light on the common weakness shared by different vision systems across various tasks. So far, adversarial training has been the most popular approach for building robust image classifiers. However, adversarial training is computationally expensive and more importantly, ill-defined for downstream tasks like image captioning. In such a scenario, mitigating the common weakness can make the task of building robust end-to-end systems tractable.
 
 In this paper, we propose \emph{Mimic and Fool}, a task agnostic adversarial attack, which exploits the non-invertibility of CNN-based \emph{feature extractors} to attack the downstream model. Given a model and its feature extractor, the proposed attack is based on the simple hypothesis that if two images are \emph{indistinguishable} for the \emph{feature extractor} then they will be \emph{indistinguishable} for the model as well. In other words, attacking the feature extractor by finding two indistinguishable images is equivalent to \emph{hacking the eyes} of the model. As an example, consider an encoder-decoder architecture like Show and Tell \cite{showandtell}, if we can successfully find two images which are mapped to the same feature by the encoder, then the two images will generate same (or similar) caption regardless of the decoder architecture. Thus to attack any model, attacking its feature extractor suffices. Based on this insight, \emph{Mimic and Fool} finds an adversarial image which can \emph{mimic} the feature of the original image thereby \emph{fooling} the model. Figure \ref{fig:main} shows examples of  \emph{Mimic and Fool} on two captioning models: Show and Tell \cite{showandtell}, Show Attend and Tell \cite{showattendandtell} and one VQA model: end-to-end neural module network (N2NMN) \cite{n2nmn}. It is crucial to note that the goal of Mimic and Fool differs from traditional adversarial attacks~\cite{43405,KurakinGB16,P18-1241,Xu-2018-CVPR}. In traditional adversarial attacks, small amount of noise is added to the image in order to fool the model to generate a different output. Whereas, in Mimic and Fool, the goal is to generate an adversarial image which can fool the model to predict the same output as the original image. As we can see from Figure \ref{fig:main}, the adversarial images obtained via \emph{Mimic and Fool} are noisy images. Such images, although noisy, pose a security risk for real-world systems. This is in line with adversarial attacks on object detectors where a large amount of noise is added~\cite{shapeshifter,Huang19}. In order to generate \emph{natural-looking} adversarial images, we also propose a modified version of our attack, namely One Image Many Outputs (OIMO). In OIMO, we start with a fixed natural image and restrict the amount of noise that can be added to the image.

Since \emph{Mimic and Fool} only requires the \emph{fine-tuned} weights of the feature extractor to attack the model, it can be thought of as a \emph{gray-box} attack. In fact, if a model does not \emph{fine-tune} its feature extractor, \emph{Mimic and Fool} can function as a black-box attack. This is because the number of possible feature extractors is limited. Hence, an adversary can generate an adversarial image per feature extractor knowing that one of these images is bound to fool the model. Furthermore, \emph{Mimic and Fool} is extremely fast and requires less computing resources since only the feature extractor needs to be loaded in the memory instead of the model. 

We perform experiments on two tasks: image captioning and visual question answering (VQA). We randomly choose $1000$ MSCOCO \cite{mscoco} validation images and study the proposed attack on three models: Show and Tell, Show Attend and Tell, and N2NMN. We get $5208$ image-question pairs from VQA v2.0 dataset \cite{balanced_vqa_v2} for the $1000$ selected images. We choose these three models since they use different feature extractors. Show and Tell uses fully connected features from Inception-v3, Show Attend and Tell uses convolutional layer features from VGG16 and N2NMN uses features from a residual network \cite{He16}. Thus the three feature extractors vary from shallow to very deep helping us to validate our proposed attack for different types of feature extractors. We consider our attack successful if the model gives the same output for original and adversarial image. 
%attack valid against invertibility.

\subsection{Contributions of this work}

The contributions of this paper are as follows: (i) We introduce the notion of a task agnostic attack. The proposed task agnostic attack, \emph{Mimic and Fool}, achieves high success rates for Show and Tell, Show Attend and Tell, N2NMN respectively. This validates our hypothesis that attacking the feature extractor suffices and also shows that the proposed attack works for different feature extractors. For image captioning models, we also compute the BLEU \cite{bleu} and METEOR \cite{meteor} score for the failure cases to show that even though the original and adversarial captions do not match exactly for these cases, they are very similar to each other. (ii) Even for \emph{One Image Many Outputs}, the proposed attack achieves decent success rate. This shows that, by adding minimal noise to the fixed image, it is possible to find an adversarial image which can mimic image feature of any arbitrary image. This result is intriguing as it suggests that the feature extractors are very chaotic in nature. (iii) Since \emph{Mimic and Fool} is task agnostic, while attacking a VQA model like N2NMN we need to run the attack for every image instead of every image-question pair. This is a huge advantage in terms of time saved for the adversary. The same will hold true for any future tasks which take multiple modalities as input with image being one of the modalities. (iv) At first glance, it seems that an invertible feature extractor will be resistant to the proposed attack. However, we show that the proposed attack also works for invertible architecture \cite{irevnet}. This shows that such architectures, despite being invertible, assign similar features to dissimilar images. Hence, invertibility is not a sufficient condition to safeguard the models against the proposed attack.  
% \textcolor{red}{We also propose a modified version of our attack to generate \emph{natural-looking} adversarial images. In the modified version, we start with a fixed natural image and restrict the amount of noise that can be added to the image. We use the same starting image for both the captioning models. Even in this restricted setting, the proposed attack achieves decent success rate.}

%The rest of the paper is organized as follows. In Section~\ref{Sect::Method}, we explain the proposed attack in detail and then we specify the training details and the hyperparameters used for the present work. In Section~\ref{Sect::Result}, we analyze the results of \emph{Mimic and Fool} on the three models. We also analyze the results of the proposed variant of \emph{Mimic and Fool} on the two captioning models; Show and Tell and Show Attend and Tell. Furthermore, we study the proposed variant on an invertible architecture. Finally in Section~\ref{Sect::Conclusion}, we give conclusions and related future work.

\section{Method}\label{Sect::Method}
\subsection{Proposed Attack}
In this section, we describe the proposed attack, \emph{Mimic and Fool}, and \emph{One Image Many Outputs} (OIMO) which is able to generate \emph{natural looking} adversarial images. Since both the attacks are task agnostic, we describe the attack in terms of the feature extractor instead of the model.

\subsubsection{\textbf{Mimic and Fool}}

Let $f: \mathbb{R}^{m\times n \times3}\longrightarrow \mathbb{R}^{d}$ denote the feature extractor of the model. Hence, $d$ will be $14\times14\times1024$ if we extract \emph{conv4} features from ResNet101 and  $d$ will be $2048$ if we use output of average pooling layer of Inception-v3 as image feature. 

Let $I_{org}\in[0,255]^{m\times n \times3}$ denote the original image. Given $I_{org}$ and a feature extractor $f$, our goal is to find an adversarial image $I_{adv}\in[0,255]^{m\times n \times3}$ which can \emph{mimic} the image features of $I_{org}$.  We model this task as a simple optimization problem given by 
%min\limits_{I} c \left\|f(I)-f(I_{org})\right\|_{2}^{2}
\begin{equation}
    \min\limits_{I}\frac{\left\|f(trunc(I))-f(I_{org})\right\|_{2}^{2}}{d}
\label{eq:maf1}
\end{equation}

where $\left\|.\right\|_{2}$ denotes $\ell_{2}-$norm and $trunc$ is truncating function which ensures that the intensity values lie in the range $[0,255]$. Although $I = I_{org}$ is a solution to the above optimization problem, it is highly unlikely that the algorithm will converge to this solution. This is because convolutional neural networks discard significant amount of spatial information as we go from lower to higher layers. Mahendran and Vedaldi\cite{MahendranV15} show that the amount of invariance increases from lower to higher layer of AlexNet and regularizers like \emph{total variation} (TV) are needed to reconstruct the original image from higher layer features of AlexNet. We start with a \emph{zero-image} and run the proposed attack for $max_{iter}$ iterations and return the final truncated image $trunc(I)$ as $I_{adv}$.

Some feature extractors such as Inception-v3 require the intensity values of the input image to be in the range $[-1,1]$. In such a case, let $I_{org}'\in[-1,1]^{m\times n \times3}$ be the scaled original image i.e. 
\begin{equation}
    I_{org}' = 2(I_{org}/255) - 1
\label{eq:maf2scaleorg}
\end{equation}
For this case, we modify the optimization problem defined in Equation \ref{eq:maf1} as follows
\begin{equation}
    \min\limits_{I}\frac{\left\|f(tanh(I))-f(I_{org}')\right\|_{2}^{2}}{d}
\label{eq:maf2}
\end{equation}

where $tanh$ ensures that the input to feature extractor lies within the required range. We run the attack for $max_{iter}$ iterations and rescale the final image $tanh(I)$ to get $I_{adv}$ i.e.
\begin{equation}
    I_{adv} = 255\left(\frac{tanh(I)+1}{2}\right)
\label{eq:maf2scaleadv}
\end{equation}

\subsubsection{\textbf{One Image Many Outputs}}
In \emph{One Image Many Outputs} (OIMO), we start with an image $I_{start}\in[0,255]^{m\times n \times3}$ instead of starting with \emph{zero-image}. The image $I_{start}$ is kept fixed throughout the experiment. In OIMO, our goal is to modify $I_{start}$ so as to \emph{mimic} the feature of $I_{org}$. Equation \ref{eq:maf1} is modified as follows

\begin{equation}
    \min\limits_{\delta}\frac{\left\|f(trunc(I_{start}+\delta))-f(I_{org})\right\|_{2}^{2}}{d}
\label{eq:oimo1}
\end{equation}

Similar to Chen \emph{et al.} \cite{P18-1241}, we modify the Equation \ref{eq:maf2} as follows

\begin{equation}
    \min\limits_{\delta}\frac{\left\|f(tanh(I_{start}''+\delta)) -f(I_{org}')\right\|_{2}^{2}}{d}
\label{eq:oimo2}
\end{equation}

where $I_{start}'' = arctanh(\lambda I_{start}')$, $I_{start}'\in[-1,1]^{m\times n \times3}$ is the scaled starting image, $\lambda$ is set to $0.9999$ to ensure invertibility of $tanh$, $\delta\in\mathbb{R}^{m\times n \times3}$ is the learnable parameter. For this attack, we reduce the value of $max_{iter}$ and initial learning rate to ensure that $I_{adv}$ looks very similar to $I_{start}$. 

%Instead of this, it is also plausible to use the clipping function $Clip_{I_{start},\epsilon}$ as in the basic iterative method of Kurakin \emph{et al.} \cite{KurakinGB16}. 

Similar to \emph{Mimic and Fool}, after running the attack for $max_{iter}$ iterations, $I_{adv}$ for Equation \ref{eq:oimo1} is $trunc(I_{start}+\delta)$. For Equation \ref{eq:oimo2}, $I_{adv}$ is given by the following equation

\begin{equation}
    I_{adv} = 255\left(\frac{tanh(I_{start}''+\delta)+1}{2}\right)
\label{eq:oimo2scaleadv}
\end{equation}
We name the proposed attack \emph{One Image Many Outputs} since all the adversarial images look very similar to $I_{start}$.

\begin{table*}[t]
\begin{center}
\renewcommand{\arraystretch}{1.2}
\begin{tabular}{c|c|c|c|c}
\hline \bf Task &\bf Model & \bf Feature Extractor & \bf Success Rate & \bf Average Time for 1000 iterations \\ \hline
\multirow{2}*{Image Captioning} &Show and Tell & Inception-v3 & 74.0 \% & 25.35 sec\\
 &Show Attend and Tell & VGG16 & 81.0 \% & 15.56 sec\\ \hline
VQA & N2NMN & ResNet-152 & 87.1 \% & 72.98 sec\\ \hline
\hline
\end{tabular}
\end{center}
\caption{Success rate of \emph{Mimic and Fool} }
\label{table:mimic-and-fool}
\end{table*}

% \multirow{2}*{Image Captioning} &Show and Tell & Inception-v3 & 74.0 \% (81.0 \%)& 25.35 sec\\
%  &Show Attend and Tell & VGG16 & 81.0 \% (87.1 \%)& 15.56 sec\\ \hline
% VQA & N2NMN & ResNet-152 & 87.1 \% & 72.98 sec\\ \hline
\subsection{Implementation Details}\label{Sect::Implementation}

As stated earlier, we study the proposed attack for two image captioning models; Show and Tell, Show Attend and Tell and one VQA model, namely, N2NMN. We train the N2NMN model on VQA v2.0 dataset for 95K iterations with expert policy followed by 65K iterations in policy search after cloning stage using the original source code\footnote{\url{https://github.com/ronghanghu/n2nmn}}. The trained N2NMN has $61.72 \%$ accuracy on VQAv2 test-dev set. For Show and Tell and Show Attend and Tell, we use already available trained models\footnote{\url{https://github.com/KranthiGV/Pretrained-Show-and-Tell-model}}\textsuperscript{,}\footnote{\url{https://github.com/DeepRNN/image_captioning}}. 

Show and Tell uses $2048$-dimensional feature from Inception-v3, Show Attend and Tell uses $14 \times 14 \times 512$ feature map from VGG16, N2NMN uses output of $res5c$ layer from ResNet-152 as image feature. The input images are of size $299\times299\times3$, $224\times224\times3$, $448\times 448\times3$ for Inception-v3, VGG16 and ResNet-152 respectively. The trained Show and Tell, Show Attend and Tell \emph{fine-tune} their respective feature extractors whereas N2NMN does not use \emph{fine-tuning}. 

For \emph{Mimic and Fool}, we set $max_{iter}$ to $1000$, $1000$ and $2000$ for Inception-v3, VGG16 and ResNet-152 respectively. The initial learning rate is set to $0.025$, $0.025$ and $0.0125$ for Inception-v3, VGG16 and ResNet-152 respectively. For \emph{One Image Many Outputs}, we set $max_{iter}$ to $300$, $500$, $500$ and set the initial learning rate to $0.0125$, $0.0125$, $0.00625$ for Inception-v3, VGG16 and ResNet-152 respectively. We use Adam \cite{adam} as the optimizer and Keras \cite{chollet2015keras} for implementing the proposed attacks. All experiments are done on a single $11$ GB GeForce GTX 1080 Ti GPU. The code for \emph{Mimic and Fool} is publicly available.\footnote{\url{https://github.com/akshay107/mimic-and-fool}}

\section{Results}\label{Sect::Result}

 For studying the two proposed attacks, $1000$ MSCOCO validation images are randomly selected. For the $1000$ selected images, there are $5208$ image-question pairs in VQA v2.0 dataset. For visual question answering, we discard those image-question pairs where the VQA model predicts the same answer for $I_{start}$ and $I_{org}$ (For Mimic and Fool, $I_{start}$ is zero-image). This is done to ensure that the VQA model predicts the same answer for $I_{start}$ and $I_{org}$ due to adversarial noise rather than language bias. The proposed attack is considered to be \emph{successful} if the model gives the same output for the original and the adversarial image. Hence for image captioning, the two captions need to be exactly the same for the attack to be successful. In the following subsections, we analyze the behavior of the two proposed attacks on the three models: N2NMN, Show and Tell and Show Attend and Tell. We also study the effectiveness of the proposed method for an invertible architecture.

\subsection{Results for Mimic and Fool}

\begin{figure}[h]
  \begin{center}
  \begin{adjustbox}{width=1\linewidth}
  \begin{tabular}{ c  c}
    \begin{minipage}{.15\textwidth}
      \includegraphics[width=\linewidth]{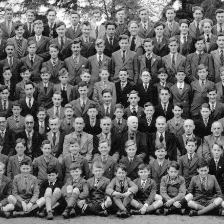}
    \end{minipage}
    & \makecell{\bf  Q: How many hands are in the picture? \\ \bf  P: 4 \\ \bf $\mathbf{P_{zero}}$: 1}\\
    \makecell{\bf Original \\ \\} & \makecell{\bf Q: What type of place is this? \\ \bf P: school \\ \bf $\mathbf{P_{zero}}$: kitchen}\\
    \begin{minipage}{.15\textwidth}
      \includegraphics[width=\linewidth]{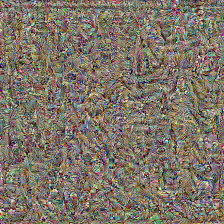}
    \end{minipage}
    & \makecell{\bf Q: Is this a recent photo? \\ \bf P: no \\ \bf $\mathbf{P_{zero}}$: yes}\\
    \bf Adversarial & \\
    \end{tabular}
  \end{adjustbox}
  \end{center}
  \caption{Example of \emph{Mimic and Fool} for N2NMN. Single adversarial image suffices for three image-question pairs. Q and P denote the question and the predicted answer respectively. $P_{zero}$ denotes the predicted answer for zero image.}
  \label{fig:table2}
\end{figure}

Table \ref{table:mimic-and-fool} shows the success rate of \emph{Mimic and Fool} for the three models. Out of $5208$ image question pairs, N2NMN predicts the same answer for $I_{org}$ and zero-image for $1707$ pairs. Out of the remaining $3501$ pairs, \emph{Mimic and Fool} is successful for $3049$ image question pairs. This yields success rate of $87.1\%$. The high success rate shows that it is possible to \emph{mimic} features extracted from a very deep network like ResNet-152 as well. Since \emph{Mimic and Fool} is \emph{task-agnostic}, we need to run the proposed attack at image level instead of image-question pair level. This is a huge advantage since it results in a drastic reduction in time. The advantage will be even more pronounced for any future tasks which have multiple modalities as input with image (or video) being one of the modalities. Figure \ref{fig:table2} shows the predicted answer by N2NMN for different image-question pairs. From Figure \ref{fig:table2}, we can see that a single adversarial image suffices for three image-question pairs. 
%\textcolor{red}{Hence, we only need to run the attack $1000$ times instead of $5208$ times.}
%For N2NMN, out of $5208$ image question pairs, \emph{Mimic and Fool} is successful for $4665$ image question pairs. This yields success rate of $89.6\%$.
% \begin{figure}[h]
% \begin{center}
%   \includegraphics[width=\linewidth]{table2.eps}
% \end{center}
%   \caption{Example of \emph{Mimic and Fool} for N2NMN. Single adversarial image suffices for three image-question pairs. Q and P denote the question and the predicted answer respectively.}
% \label{fig:table2}
% \end{figure}

\begin{figure*}[htbp]
  \begin{center}
  \begin{adjustbox}{width=1\linewidth}
  \Large
  \begin{tabular}{ c  c  c  c  c  c}
     \multicolumn{6}{c}{\textbf{Show and Tell}} \\
    \rule{0pt}{10ex}
    \begin{minipage}{.25\textwidth}
      \includegraphics[width=\linewidth]{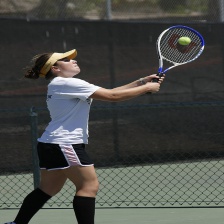}
    \end{minipage}
    & \begin{minipage}{.25\textwidth}
      \includegraphics[width=\linewidth]{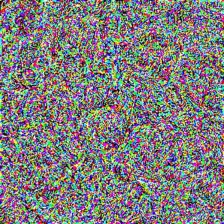}
    \end{minipage} & 
    \begin{minipage}{.25\textwidth}
      \includegraphics[width=\linewidth]{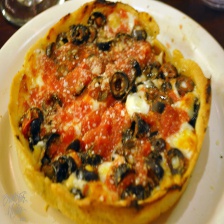}
    \end{minipage} & 
    \begin{minipage}{.25\textwidth}
      \includegraphics[width=\linewidth]{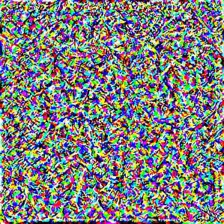}
    \end{minipage} & 
    \begin{minipage}{.25\textwidth}
      \includegraphics[width=\linewidth]{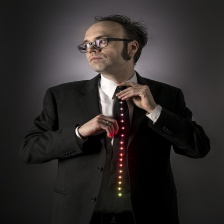}
    \end{minipage} & 
    \begin{minipage}{.25\textwidth}
      \includegraphics[width=\linewidth]{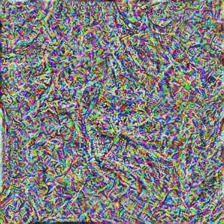}
    \end{minipage} \\ 
    \makecell{\bf a female tennis player \\ \bf in action on the court.} &
    \makecell{\bf a female tennis player \\ \bf in action on the court.} &
    \makecell{\bf a pizza sitting in top \\ \bf of a white plate.} &
    \makecell{\bf a pizza sitting in top \\ \bf of a white plate.} &
    \makecell{\bf \textit{a man in a suit and tie} \\ \bf \textit{standing in the room.}} &
    \makecell{\bf \textit{a man in a suit and tie} \\ \bf \textit{is smiling.}} \\[0.4cm]
    \multicolumn{6}{c}{\textbf{Show Attend and Tell}} \\
    \rule{0pt}{10ex}
    \begin{minipage}{.25\textwidth}
      \includegraphics[width=\linewidth]{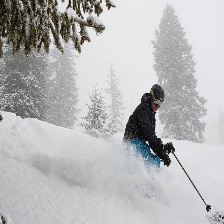}
    \end{minipage}
    & \begin{minipage}{.25\textwidth}
      \includegraphics[width=\linewidth]{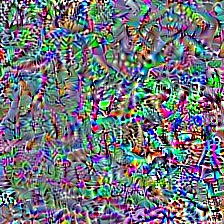}
    \end{minipage} & 
    \begin{minipage}{.25\textwidth}
      \includegraphics[width=\linewidth]{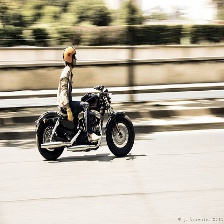}
    \end{minipage} & 
    \begin{minipage}{.25\textwidth}
      \includegraphics[width=\linewidth]{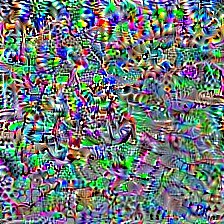}
    \end{minipage} & 
    \begin{minipage}{.25\textwidth}
      \includegraphics[width=\linewidth]{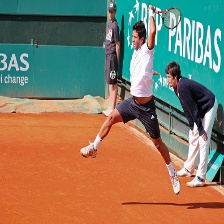}
    \end{minipage} & 
    \begin{minipage}{.25\textwidth}
      \includegraphics[width=\linewidth]{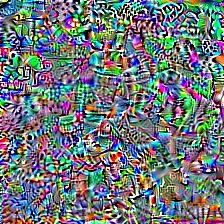}
    \end{minipage} \\
    \makecell{\bf a man is riding skis  \\ \bf down a snow covered\\ \bf slope.} &
    \makecell{\bf a man is riding skis  \\ \bf down a snow covered\\ \bf slope.} &
    \makecell{\bf a man riding a \\ \bf motorcycle down a\\ \bf street.} &
    \makecell{\bf a man riding a \\ \bf motorcycle down a\\ \bf street.} &
    \makecell{\bf \textit{a man swinging a}  \\ \bf \textit{tennis racquet on} \\ \bf \textit{a tennis court.}} &
    \makecell{\bf \textit{a man is playing}  \\ \bf \textit{tennis on a} \\ \bf \textit{tennis court.}} \\   
  \end{tabular}
  \end{adjustbox}
  \end{center}
  \caption{Examples of \emph{Mimic and Fool}. For both the captioning models, the figure shows two successful and one unsuccessful original and adversarial images along with the predicted captions. Unsuccessful cases are shown in italics.}
  \label{fig:maf}
\end{figure*}

As we can see from Table \ref{table:mimic-and-fool}, \emph{Mimic and Fool} is very fast. The attack only takes around $25$ seconds for generating adversarial images for Show and Tell. The time taken for Show, Attend and Tell is even less since VGG16 is a shallower network. The proposed attack achieves success rate of $74.0\%$ and $81.0\%$ for Show and Tell and Show Attend and Tell respectively. This is especially encouraging result since generating exactly the same caption for an adversarial image is a very challenging task. This is because, as observed by Chen \emph{et al.} \cite{P18-1241}, the number of possible captions are infinite which makes a captioning system harder to attack than an image classifier. Our results show that in order to generate the same caption, it suffices to attack just the encoder of the captioning model.  This validates our initial hypothesis that in order to attack any model, attacking its feature extractor suffices. For the unsuccessful cases, the predicted captions for original and adversarial images are very similar. Figure \ref{fig:maf} shows two successful and one unsuccessful examples of \emph{Mimic and Fool} for Show and Tell and Show Attend and Tell. As we can see from Figure \ref{fig:maf} that for the unsuccessful cases, the predicted captions for the original and adversarial images have a large amount of overlap. We also calculate the BLEU and METEOR score, using the pipeline provided by Sharma \emph{et al.} \cite{sharma2017nlgeval}, for unsuccessful adversarial cases as shown in Table \ref{table:main-result}. We use the predicted caption for the original image as reference while calculating these metrics.

\begin{table*}[ht]
\begin{center}
\renewcommand{\arraystretch}{1.2}
\begin{tabular}{c|c|c|c|c|c|c}
\hline \bf Model \bf & \bf Attack & \bf BLEU-1 & \bf BLEU-2 & \bf BLEU-3 & \bf BLEU-4 & \bf METEOR\\ \hline
Show and Tell & \begin{tabular}[c]{@{}c@{}}Show-and-Fool~\cite{P18-1241} \\ Mimic and Fool\\ OIMO\end{tabular} & \begin{tabular}[c]{@{}l@{}}$0.560$\\$0.597$\\ $0.593$\end{tabular} &
\begin{tabular}[c]{@{}l@{}}$0.394$\\$0.464$\\ $0.459$\end{tabular} &
\begin{tabular}[c]{@{}l@{}}$0.266$\\$0.348$\\ $0.350$\end{tabular} &
\begin{tabular}[c]{@{}l@{}}$0.205$\\$0.264$\\ $0.270$\end{tabular} &
\begin{tabular}[c]{@{}l@{}}$0.301$\\$0.320$\\ $0.322$\end{tabular} \\ 
\hline
Show Attend and Tell & \begin{tabular}[c]{@{}c@{}}EM~\cite{yan2019attack}\\ SSVM~\cite{yan2019attack}\\Mimic and Fool\\ OIMO\end{tabular} & 
\begin{tabular}[c]{@{}l@{}}$0.765$\\$0.635$\\$0.639$\\$0.594$\end{tabular} &
\begin{tabular}[c]{@{}l@{}}$0.650$\\$0.501$\\$0.530$\\$0.468$\end{tabular} &
\begin{tabular}[c]{@{}l@{}}$0.529$\\$0.409$\\$0.421$\\$0.359$\end{tabular} &
\begin{tabular}[c]{@{}l@{}}$0.423$\\$0.300$\\$0.333$\\$0.284$\end{tabular} &
\begin{tabular}[c]{@{}l@{}}$0.425$\\$0.337$\\$0.368$\\$0.336$\end{tabular} \\ 
\hline
\end{tabular}
\end{center}
\caption{BLEU and METEOR scores for \textbf{unsuccessful} cases. OIMO refers to \emph{One Image Many Outputs}.}
\label{table:main-result}
\end{table*}

\subsection{Results for One Image Many Outputs}

The main idea behind \emph{One Image Many Outputs} is to generate natural-looking adversarial images. We randomly choose an image from MSCOCO training set as the starting image. Figure \ref{fig:I_start} shows the starting image ($I_{start}$) for \emph{One Image Many Outputs} along with the predicted captions of Show And Tell and Show Attend and Tell. We use the same $I_{start}$ for N2NMN. Similar to Mimic and Fool, we discard $1713$ image-question pairs for which N2NMN predicts the same answer for $I_{org}$ and $I_{start}$.

% \begin{figure}[h]
% \begin{center}
%   \includegraphics[width=\linewidth]{I_start.eps}
% \end{center}
%   \caption{$I_{start}$ for \emph{One Image Many Outputs} and the predicted captions.}
% \label{fig:I_start}
% \end{figure}

\begin{figure}[h]
  \begin{center}
  \begin{adjustbox}{width=1\linewidth}
  \begin{tabular}{ c  c}
    \begin{minipage}{.22\textwidth}
      \includegraphics[width=\linewidth]{I_start_img.png}
    \end{minipage}
    & \makecell{\bf Show and Tell: a plastic container \\ \bf filled with lots of food. \\ \\ \\ \\ \bf Show Attend and Tell: a tray filled \\ \bf with different types of food.}\\
    \end{tabular}
  \end{adjustbox}
  \end{center}
  \caption{ $I_{start}$ for \emph{One Image Many Outputs} and the predicted captions.}
\label{fig:I_start}
\end{figure}

\begin{figure}[h!]
  \begin{center}
  \begin{adjustbox}{width=1\linewidth}
  \begin{tabular}{ c  c}
    \begin{minipage}{.15\textwidth}
      \includegraphics[width=\linewidth]{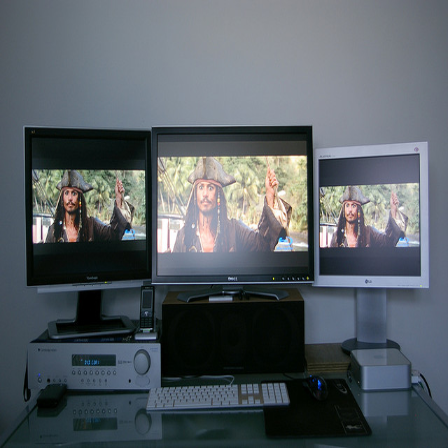}
    \end{minipage}
    & \makecell{\bf Q: Is there a thriller playing on the screen? \\ \bf P: no \\ \bf $\mathbf{P_{I_{start}}}$: yes}\\
    \makecell{\bf Original \\ \\} & \makecell{\bf Q: Is this person sick? \\ \bf P: no \\ \bf $\mathbf{P_{I_{start}}}$: yes}\\
    \begin{minipage}{.15\textwidth}
      \includegraphics[width=\linewidth]{oimo_vqa_adv_img.jpg}
    \end{minipage}
    & \makecell{\bf Q: Is any one of these a TV? \\ \bf P: yes \\ \bf $\mathbf{P_{I_{start}}}$: no}\\
    \bf Adversarial & \\
    \end{tabular}
  \end{adjustbox}
  \end{center}
  \caption{Example of \emph{One Image Many Outputs} for N2NMN. Single adversarial image suffices for three image-question pairs. Q and P denote the question and the predicted answer respectively. $P_{I_{start}}$ denotes the predicted answer for $I_{start}$.}
  \label{fig:oimo-vqa}
\end{figure}

% \begin{figure*}[t]
% \begin{center}
%   \includegraphics[width=\linewidth]{maf.eps}
% \end{center}
%   \caption{Examples of \emph{Mimic and Fool}. For both the captioning models, the figure shows two successful and one unsuccessful original and adversarial image along with the predicted captions.}
% \label{fig:maf}
% \end{figure*}

% \begin{figure*}[h!]
% \begin{center}
%   \includegraphics[width=\linewidth]{oimo.eps}
% \end{center}
%   \caption{Examples of \emph{One Image Many Outputs}. For both the captioning models, the figure shows two successful and one unsuccessful original and adversarial image along with the predicted captions. For adversarial images, the captions within brackets are predicted by the other captioning model i.e. for Show and Tell, the captions within brackets are captions predicted by Show Attend and Tell and vice-versa.}
% \label{fig:oimo}
% \end{figure*}

\begin{figure*}[h!]
  \begin{center}
  \begin{adjustbox}{width=1\linewidth}
  \Large
  \begin{tabular}{ c  c  c  c  c  c}
     \multicolumn{6}{c}{\textbf{Show and Tell}} \\
    \rule{0pt}{10ex}
    \begin{minipage}{.25\textwidth}
      \includegraphics[width=\linewidth]{st_oimc_org_img1.jpg}
    \end{minipage}
    & \begin{minipage}{.25\textwidth}
      \includegraphics[width=\linewidth]{st_oimc_adv_img1.jpg}
    \end{minipage} & 
    \begin{minipage}{.25\textwidth}
      \includegraphics[width=\linewidth]{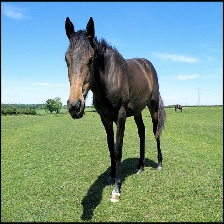}
    \end{minipage} & 
    \begin{minipage}{.25\textwidth}
      \includegraphics[width=\linewidth]{st_oimc_adv_img2.jpg}
    \end{minipage} & 
    \begin{minipage}{.25\textwidth}
      \includegraphics[width=\linewidth]{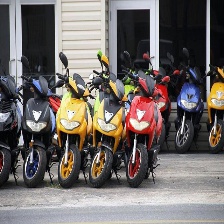}
    \end{minipage} & 
    \begin{minipage}{.25\textwidth}
      \includegraphics[width=\linewidth]{st_oimc_adv_img3.jpg}
    \end{minipage} \\ 
    \makecell{\bf a woman standing in \\ \bf front of a \\ \bf refrigerator.\\ \\ \\ } &
    \makecell{\bf ST: a woman standing \\ \bf in front of a \\ \bf refrigerator. \\ \bf SAT: a close up of \\ \bf a tray of food.} &
    \makecell{\bf a brown horse \\ \bf standing on top of \\ \bf a lush green field.\\ \\ \\ } &
    \makecell{\bf ST: a brown horse \\ \bf standing on top of \\ \bf a lush green field.\\ \bf SAT: a close up of \\ \bf a tray of food.} &
    \makecell{\bf \textit{a row of motorcycles} \\ \bf \textit{parked next to each}\\ \bf \textit{other.}\\ \\ \\} &
    \makecell{\bf \textit{ST: a row of parked} \\ \bf \textit{motorcycles sitting} \\ \bf \textit{next to each other.}\\ \bf \textit{SAT: a close up of} \\ \bf \textit{a tray of food.}} \\[0.4cm]
    \multicolumn{6}{c}{\textbf{Show Attend and Tell}} \\
    \rule{0pt}{10ex}
    \begin{minipage}{.25\textwidth}
      \includegraphics[width=\linewidth]{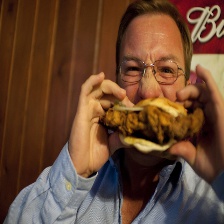}
    \end{minipage}
    & \begin{minipage}{.25\textwidth}
      \includegraphics[width=\linewidth]{sat_oimc_adv_img1.jpg}
    \end{minipage} & 
    \begin{minipage}{.25\textwidth}
      \includegraphics[width=\linewidth]{sat_oimc_org_img2.jpg}
    \end{minipage} & 
    \begin{minipage}{.25\textwidth}
      \includegraphics[width=\linewidth]{sat_oimc_adv_img2.jpg}
    \end{minipage} & 
    \begin{minipage}{.25\textwidth}
      \includegraphics[width=\linewidth]{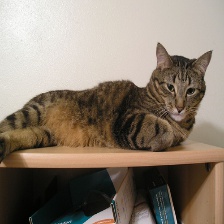}
    \end{minipage} & 
    \begin{minipage}{.25\textwidth}
      \includegraphics[width=\linewidth]{sat_oimc_adv_img3.jpg}
    \end{minipage} \\
    \makecell{\bf a man holding a hot \\ \bf dog in his hand.\\ \\ \\ \\ \\} &
    \makecell{\bf SAT: a man holding a  \\ \bf hot dog in his hand.\\ \bf ST: a bunch of \\ \bf different types of \\ \bf food on a table.\\ \\} &
    \makecell{\bf a man holding a  \\ \bf tennis racquet on a  \\ \bf tennis court.\\ \\ \\ \\ } &
    \makecell{\bf SAT: a man holding a  \\ \bf tennis racquet on a \\ \bf tennis court.\\ \bf ST: a table topped  \\ \bf with lots of different \\ \bf types of vegetables.} &
    \makecell{\bf \textit{a cat laying on top of}  \\ \bf \textit{a wooden desk.}\\ \\ \\ \\ \\ } &
    \makecell{\bf \textit{SAT: a cat laying on}   \\ \bf \textit{top of a desk.}\\ \bf \textit{ST: a lunch box} \\ \bf \textit{with a variety of} \\ \bf \textit{vegetables.}\\ \\} 
  \end{tabular}
  \end{adjustbox}
  \end{center}
  \caption{Examples of \emph{One Image Many Outputs}. For both the captioning models, the figure shows two successful and one unsuccessful original and adversarial images along with the predicted captions. Unsuccessful cases are shown in italics. For adversarial images, ST and SAT denote Show and Tell and Show Attend and Tell respectively.}
  \label{fig:oimo}
\end{figure*}

In \emph{One Image Many Outputs}, we reduce the value of $max_{iter}$ and the initial learning rate to ensure that the adversarial image $I_{adv}$ looks very similar to $I_{start}$. Reduction in $max_{iter}$ results in even faster running time than \emph{Mimic and Fool}. Table \ref{table:oimo} shows the success rate of \emph{One Image Many Outputs} for Show and Tell, Show Attend and Tell and N2NMN. As we can see from Table \ref{table:mimic-and-fool} and Table \ref{table:oimo}, the success rate reduces for \emph{One Image Many Outputs} in comparison to \emph{Mimic and Fool}. This is intuitive since in \emph{One Image Many Outputs}, the reduced value of $max_{iter}$ and initial learning rate allows for less adversarial noise. Figure~\ref{fig:oimo-vqa} shows an example of OIMO for N2NMN. Similar to Mimic and Fool, a single adversarial image suffices for multiple image-question pairs.
% \textcolor{red}{For both the captioning models, \emph{One Image Many Outputs} takes under $10$ seconds per image.}
\begin{table}[h]
\begin{center}
\renewcommand{\arraystretch}{1.2}
\begin{tabular}{c|c|c}
\hline \bf Model & \bf Success Rate & \bf Time (in sec.)\\ \hline
Show and Tell & 56.9 \%  & 7.61\\
Show Attend and Tell & 50.3 \%  & 7.78\\
N2NMN & 72.8 \% & 36.50\\
\hline
\end{tabular}
\end{center}
\caption{Success rate of \emph{One Image Many Outputs} }
\label{table:oimo}
\end{table}

% Show and Tell & 56.9 \% (67.2 \%) & 7.61\\
% Show Attend and Tell & 50.3 \% (58.6 \%) & 7.78\\
% N2NMN & 72.8 \% & 36.50\\

\begin{table*}[h]
\begin{center}
\renewcommand{\arraystretch}{1.2}
\begin{tabular}{c|c|c|c|c}
\hline \bf Task & \bf Model & \bf Method &\bf Success Rate  & \bf Time (in sec)\\ \hline
\multirow{3}*{Image Captioning} & Show and Tell & Show-and-Fool~\cite{P18-1241} & 95.1 \% & 177.93 \\ \cline{2-5}
                                &\multirow{2}*{Show Attend and Tell} & EM~\cite{yan2019attack}   & 77.1 \% & 20.69 \\
                                &     & SSVM~\cite{yan2019attack} & 82.1 \% & 18.73 \\
                                 \hline
VQA & N2NMN & Xu et al.~\cite{Xu-2018-CVPR} & 100.0 \% & $8.77\times n$ \\
\hline
\end{tabular}
\end{center}
\caption{Success rate and Time for task-specific methods. }
\label{table:task-specific}
\end{table*}

From Table \ref{table:oimo}, we can see that \emph{One Image Many Outputs} takes under $8$ seconds per image for both the captioning models. Considering this reduction and the fact that the attack is successful only when there is an exact match of captions, the success rate of \emph{One Image Many Outputs} is impressive. Similar to \emph{Mimic and Fool}, we find that for the unsuccessful cases of \emph{One Image Many Outputs}, the captions predicted by the model for the adversarial and original images are very similar to each other. Table \ref{table:main-result} shows the BLEU and METEOR score for the unsuccessful cases of \emph{One Image Many Outputs}. This result shows that even when $I_{adv}$ is very similar to $I_{start}$, it can \emph{mimic} features of an arbitrary image. This shows that CNN-based feature extractors are chaotic in nature.

Figure \ref{fig:oimo} shows two successful and one unsuccessful examples (shown in italics) of \emph{One Image Many Outputs} for Show and Tell and Show Attend and Tell. For the adversarial images in Figure \ref{fig:oimo}, ST and SAT denote Show and Tell and Show Attend and Tell respectively. As we can see from Figure \ref{fig:oimo}, all the six adversarial images are very similar to the starting image, $I_{start}$. Also for the unsuccessful cases, the original and adversarial captions have a large amount of overlap and are semantically similar. 
%\textcolor{red}{the captions within brackets are the predicted captions by the other captioning model i.e. the captioning model which is not under attack.}
In Figure \ref{fig:oimo}, we see that for Show and Tell, the captions predicted by Show Attend and Tell for the three adversarial images are the same. Similarly for Show Attend and Tell, although the captions predicted by Show and Tell are different, they are semantically similar. Moreover, for both the captioning models, the predicted captions by the other captioning model are relevant captions for the starting image, $I_{start}$. In fact, we find that when the $1000$ adversarial images for Show And Tell are given as input to Show Attend and Tell, there are only $15$ unique captions. All these $15$ captions are relevant captions for $I_{start}$. Similarly, when the $1000$ adversarial images for Show Attend and Tell are given as input to Show and Tell, there are only $82$ unique captions, most of which are relevant to $I_{start}$. We find that Show and Tell generates irrelevant captions for $I_{start}$ only for $32$ out of $1000$ adversarial images. Since the two captioning models use different feature extractors, this result shows that the proposed attack is very dependent on the feature extractor. In other words, ensuring that the two images are \emph{indistinguishable} for one feature extractor does not ensure that they will be \emph{indistinguishable} for another feature extractor. More examples of the two proposed attacks are provided in the supplementary material.\footnote{\url{https://www.isical.ac.in/~utpal/resources.php}}

\subsection{Comparison with task specific attack}
In this section, we compare our proposed attack, OIMO with other task-specific attacks. For Show and Tell, we use Show-and-Fool~\cite{P18-1241}. For Show Attend and Tell, we use EM and SSVM methods of Yan \emph{et al.}~\cite{yan2019attack}. For N2NMN, we use the VQA attack of Xu \emph{et al.}~\cite{Xu-2018-CVPR}. For Show-and-Fool and EM and SSVM methods, we use the official implementation.\footnote{\url{https://github.com/IBM/Image-Captioning-Attack}}\footnote{\url{https://github.com/wubaoyuan/adversarial-attack-to-caption}} We implement the attack proposed by Xu et al.~\cite{Xu-2018-CVPR} using the default parameters mentioned in the paper. Similar to OIMO, we start with  $I_{start}$ and run the task specific attacks in order to generate adversarial outputs. Table~\ref{table:task-specific} shows the success rate and time for different task-specific methods. Show-and-Fool achieves a success rate of $95.1 \%$ and takes $177.93$ seconds per image. The EM and SSVM take less time for Show Attend and Tell but have lower success rates. In contrast, OIMO takes around $8$ seconds per image for both the captioning models. 
%\textcolor{red}{around $190$}
%\textcolor{red}{Although our proposed attack achieves a lower success rate, we can see from Table~\ref{table:main-result} that for unsuccessful cases, OIMO achieves higher BLEU and METEOR score than Show-and-Fool.}
For unsuccessful cases, like OIMO, Show-and-Fool and EM and SSVM generate similar captions for original and adversarial images as evident from high BLEU and METEOR scores in Table~\ref{table:main-result}. We find that for the adversarial images generated by Show-and-Fool, Show Attend and Tell generates only $11$ unique captions, all of which are relevant captions for $I_{start}$. Chen \emph{et al.}~\cite{P18-1241} study the transferability of Show-and-Fool between the captioning models, however in their study, the two captioning models use the same feature extractor. Similarly, we obtain only 3 and 5 unique captions from Show and Tell for adversarial images of EM and SSVM respectively. All these captions are relevant for $I_{start}$. Xu et al.~\cite{Xu-2018-CVPR} achieve $100.0\%$ success rate. The attack takes $8.77$ seconds for each image-question pair. The factor $n$ in the time for Xu et al. in Table~\ref{table:task-specific} signifies the average number of questions per image, which can be arbitrarily large.

\subsection{OIMO for invertible architecture}
\begin{figure}[h]
\begin{center}
   \includegraphics[width=0.7\linewidth]{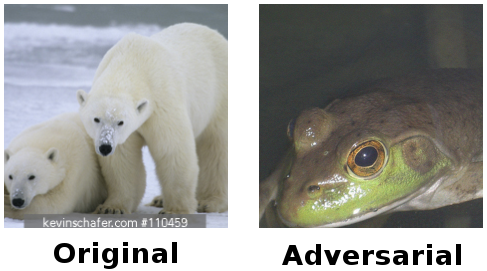}
\end{center}
   \caption{Both the images are classified as \emph{ice bear} by bijective i-RevNet.}
\label{fig:irev}
\end{figure}

Recently, Jacobsen \emph{et al.} \cite{irevnet} propose a deep invertible architecture, i-RevNet which learns a one-to-one mapping between image and its feature. These networks achieve impressive accuracy on ILSVRC-2012 \cite{ILSVRC15}. For experimentation, we choose bijective i-RevNet which takes images of size $224\times224\times3$ as input and the corresponding feature is of size $3072\times7\times7$. We use the pretrained i-RevNet provided in the official implementation\footnote{\url{https://github.com/jhjacobsen/pytorch-i-revnet}} to test our proposed attack, \emph{One Image Many Outputs}. We randomly choose $100$ correctly classified images belonging to $41$ different classes from the validation set of ILSVRC-2012. Furthermore, we choose a starting image, $I_{start}$, belonging to a different class. We also restrict the search space for adversarial images using the clipping function $Clip_{I_{start},\epsilon}$ (i.e. the adversarial noise is clipped to ensure that the adversarial image $I_{adv}$ will lie in an $\epsilon$ $\ell_{\infty}$-neighborhood of $I_{start}$). Starting with $I_{start}\in[0,255]^{224\times 224 \times3}$, we run the proposed attack, OIMO, in order to \emph{mimic} the feature for $100$ images. Table \ref{table:oimo-irev} shows the success rate for different values of $\epsilon$. The high success rate shows that the proposed attack can be applied for invertible architecture like i-RevNet as well. This is because i-RevNet, despite being invertible, assigns similar features to dissimilar images. Figure \ref{fig:irev} shows one such successful adversarial example. 

\begin{table}[h]
\begin{center}
\renewcommand{\arraystretch}{1.2}
\begin{tabular}{c|c}
\hline $\bm{\epsilon}$ & \bf Success Rate  \\ \hline
2 & 86.0 \% \\
5 & 99.0 \% \\
10 & 100.0 \% \\
\hline
\end{tabular}
\end{center}
\caption{Success rate of \emph{One Image Many Outputs} for i-RevNet}
\label{table:oimo-irev}
\end{table}

\section{Quantitative study of Adversarial Noise}

Table~\ref{table:psnr} shows the peak signal-to-noise ratio (PSNR) for OIMO and task-specific methods. The PSNR is calculated as follows
\begin{equation}
\begin{split}
    PSNR = 20\:log_{10}\left(\frac{255.0}{\sqrt{MSE}}\right)\\ 
    \text{ where } MSE = \frac{\left\|I_{adv}-I_{start}\right\|_{2}^{2}}{m\times n \times 3}
\end{split}
\end{equation}
where $I_{adv}, I_{start} \in [0,255]^{m\times n \times 3}.$

From Table~\ref{table:psnr}, it is evident that the PSNR is low for OIMO in comparison with other task-specific methods. This is mainly because task-specific methods can exploit the deficiencies of encoder as well as the decoder and such attack methods can be stopped at the exact instant when an adversarial image leads to the desired output. Agnosticity, in any form, generally leads to more noise. As an example, \emph{image-agnostic} universal adversarial perturabations (UAP)~\cite{uap} are quasi-perceptible instead of being imperceptible.

\begin{table}[h]
\begin{center}
\renewcommand{\arraystretch}{1.2}
\begin{tabular}{c|c|c}
\hline \bf Model & \bf Attack & \bf PSNR (mean $\pm$ std)  \\ \hline
\multirow{2}*{Show and Tell} & Show-and-Fool~\cite{P18-1241} & 52.5 $\pm$ 6.7 \\
                             & OIMO & 23.8 $\pm$ 0.6 \\ \hline
                             
\multirow{3}*{Show Attend and Tell} & SSVM~\cite{yan2019attack} & 42.1 $\pm$ 1.2 \\
                                    & EM~\cite{yan2019attack}   & 40.4 $\pm$ 0.9 \\
                                    & OIMO & 26.1 $\pm$ 1.1 \\ \hline
\multirow{2}*{N2NMN}                & Xu et al.~\cite{Xu-2018-CVPR} & 33.8 $\pm$ 3.7 \\
                                    & OIMO & 27.6 $\pm$ 0.5\\
\hline
\end{tabular}
\end{center}
\caption{PSNR between $I_{adv}$ and $I_{start}$ for \emph{One Image Many Outputs} (OIMO) and task-specific methods.}
\label{table:psnr}
\end{table}

\begin{table}[h]
\begin{center}
\renewcommand{\arraystretch}{1.2}
\begin{tabular}{c|c|c}
\hline \bf Model & \bf Attack & \bf SSIM (mean $\pm$ std)  \\ \hline
\multirow{2}*{Show and Tell} & MAF & $1.8 \times 10^{-4} \pm 1.3 \times 10^{-3}$ \\
                             & OIMO & $6.1 \times 10^{-4} \pm 2.9 \times 10^{-3}$ \\ \hline
\multirow{2}*{Show Attend and Tell} & MAF &  $7.5 \times 10^{-4} \pm 2.7 \times 10^{-3}$ \\
                                    & OIMO & $6.8 \times 10^{-4} \pm 4.2 \times 10^{-3}$ \\ \hline
\multirow{2}*{N2NMN} & MAF & $5.6 \times 10^{-4} \pm 1.5 \times 10^{-3}$ \\
                     & OIMO & $4.5 \times 10^{-4}$ $\pm$ $2.2 \times 10^{-3}$ \\
\hline
\end{tabular}
\end{center}
\caption{SSIM between $I_{adv}$ and $I_{org}$ for \emph{Mimic and Fool} (MaF) and \emph{One Image Many Outputs} (OIMO).}
\label{table:ssim}
\end{table}

Table~\ref{table:ssim} shows the SSIM~\cite{ssim} values between $I_{adv}$ and $I_{org}$ for the proposed methods. The \emph{near-zero} values of SSIM clearly show that there is no resemblance between the original and adversarial image.

\section{Conclusion and Future Work}\label{Sect::Conclusion}

In this paper, we proposed a task agnostic adversarial attack, \emph{Mimic and Fool}. The proposed attack exploits the non-invertibility of CNN-based feature extractors and is based on the hypothesis that if two images are \emph{indistinguishable} for the \emph{feature extractor} then they will be \emph{indistinguishable} for the model as well. The high success rate of \emph{Mimic and Fool} for three models across two tasks validates this hypothesis. We also show that the proposed attack works regardless of the depth of the feature extractor. Due to the task-agnostic nature, we need to run the attack only at image-level which is a huge advantage in terms of time saved for tasks involving multiple modalities as input. We further propose a variant of \emph{Mimic and Fool}, named \emph{One Image Many Outputs}, which generates \emph{natural-looking} adversarial images. The results for this variant of the attack show that it is possible to \emph{mimic} features of an arbitrary image by making minimal changes to a fixed image. This is an important insight into the nature of CNN-based feature extractors. We also demonstrate the applicability of the proposed attack for invertible architectures like i-RevNet.

As part of future work, from an attack perspective, one can explore different task-agnostic strategies which will work successfully with just the \emph{pretrained} weights of the feature extractor. We found that using \emph{pretrained} instead of \emph{fine-tuned} weights leads to drop in success rate of the proposed attack. From defense perspective, we show that invertible architectures like iRevNet are not robust to the proposed attack. Hence, one can explore different feature extractors which are resistant to the proposed attack. If successful, one can use these feature extractors to develop end-to-end systems and check their robustness to task-agnostic as well as task-specific attacks.

% \bibliographystyle{IEEEtran}
% \bibliography{references}

\end{document}